%% file: NECO-03-17-2829R1-Source.tex
\documentclass[12pt]{article}
\usepackage{amsmath}
\usepackage{times}
\usepackage{graphicx}
\usepackage{color}
\usepackage{multirow}
\usepackage{subcaption}
\usepackage{url}
\usepackage[authoryear]{natbib}
\newcommand{\parencite}{\citep}
\newcommand{\textcite}{\citet}
\renewcommand{\cite}{\citealt}
\usepackage{booktabs}       
\usepackage{amsfonts}       
\usepackage{nicefrac}       
\usepackage{microtype}      
\usepackage{amssymb}
\usepackage{units}
\usepackage{graphics}
\usepackage{graphicx}
\usepackage{floatrow} 
\usepackage{color}   
\usepackage{alltt}
\usepackage{amsmath}
\usepackage{amsfonts}
\usepackage{adjustbox}
\usepackage{todonotes}
\usepackage{subcaption}
\usepackage{rotating}
\usepackage{bbm}
\usepackage{latexsym}
\usepackage{fullpage}
\usepackage{color}
\usepackage[linesnumbered,ruled,vlined]{algorithm2e}
\usepackage{array}
\usepackage{enumitem}

\newcolumntype{$}{>{\global\let\currentrowstyle\relax}}
\newcolumntype{^}{>{\currentrowstyle}}

\newcommand{\captionfonts}{\normalsize}

\makeatletter  
\long\def\@makecaption#1#2{%
  \vskip\abovecaptionskip
  \sbox\@tempboxa{{\captionfonts #1: #2}}%
  \ifdim \wd\@tempboxa >\hsize
    {\captionfonts #1: #2\par}
  \else
    \hbox to\hsize{\hfil\box\@tempboxa\hfil}%
  \fi
  \vskip\belowcaptionskip}
\makeatother   

\DeclareCaptionFont{xipt}{\fontsize{12}{12}\mdseries}
\usepackage[font=xipt,labelfont=bf]{caption}

\begin{document}
\hspace{13.9cm}

\ \vspace{20mm}\\
{\LARGE Learning Simpler Language Models \\
with the Differential State Framework}
\ \\
{\bf \large
Alexander G. Ororbia II$^{\displaystyle 1}$} \\
{\bf \large
Tomas Mikolov$^{\displaystyle 2}$} \\
{\bf \large
David Reitter$^{\displaystyle 1}$} \\
{$^{\displaystyle 1}$The Pennsylvania State University}\\
  {$^{\displaystyle 2}$Facebook}\\


\thispagestyle{empty}

\begin{abstract}
Learning useful information across long time lags is a critical and difficult problem for temporal neural models in tasks such as language modeling. Existing architectures that address the issue are often complex and costly to train.  The Differential State Framework (DSF) is a simple and high-performing design that unifies previously introduced gated neural models. 
DSF models maintain longer-term memory by learning to interpolate between a fast-changing data-driven representation and a slowly changing, implicitly stable state. This requires hardly any more parameters than a classical, simple recurrent network.
Within the DSF framework, a new architecture is presented, the Delta-RNN.  In language modeling at the word and character levels, the Delta-RNN outperforms popular complex architectures, such as the Long Short Term Memory (LSTM) and the Gated Recurrent Unit (GRU), and, when regularized, performs comparably to several state-of-the-art baselines. At the subword level, the Delta-RNN's performance is comparable to that of complex gated architectures. 
\end{abstract}
\section{Introduction}
\label{intro}

Recurrent neural networks are increasingly popular models for sequential data. The simple recurrent neural network (RNN) architecture \parencite{elman1990finding} is, however, not suitable for capturing longer-distance dependencies.  Architectures that address this shortcoming include the Long Short-Term Memory (LSTM, \cite{hochreiter1997long}), the Gated Recurrent Unit (GRU,  \cite{chung2014empirical,chung2015gated}), and the structurally constrained recurrent network (SCRN, \cite{mikolov2014learning}). While these can capture some longer-term patterns (20 to 50 words), their structural complexity makes it difficult to understand what is going on inside. One exception is the SCRN architecture, which is by design simple to understand.  It shows that the memory acquired by complex LSTM models on language tasks does correlate strongly with simple weighted bags-of-words. This demystifies the abilities of the LSTM model to a degree: while some authors have suggested that the LSTM understands the language and even the thoughts being expressed in sentences \parencite{choudhury_thought_2015}, it is arguable whether this could be said about a model that performs equally well and is based on representations that are essentially equivalent to a bag of words.

One property of recurrent architectures that allows for the formation of longer-term memory is the self-connectedness of the basic units: this is most explicitly shown in the SCRN model, where one hidden layer contains neurons that do not have other recurrent connections except to themselves. Still, this architecture has several drawbacks: one has to choose the size of the fully connected and self-connected recurrent layers, and the model is not capable of modeling non-linearities in the longer-term memory component.

In this work, we aim to increase representational efficiency, i.e., the ratio of performance to acquired parameters.  We simplify the model architecture further and develop several variants under the \emph{Differential State Framework}, where the hidden layer state of the next time step is a function of its current state and the delta change computed by the model. 
We do not present the Differential State Framework as a model of human memory for language. However, we point out its conceptual origins in Surprisal Theory \parencite{boston2008parsingcosts,hale2001surprisal,levy2008expectation}, which posits that the human language processor develops complex expectations of future words, phrases, and syntactic choices, and that these expectations and deviations from them (surprisal) guide language processing, e.g., in reading comprehension.  How complex the models are (in the human language processor) that form the expectation is an open question. The cognitive literature has approached this with existing parsing algorithms, probabilistic context-free grammars, or n-gram language models.  We take a connectionist perspective.  The Differential State Framework proposes to not just generatively develop expectations and compare them with actual state changes caused by observing new input;
it explicitly maintains \emph{gates} as a form of high-level error correction and interpolation.  An instantiation, the Delta-RNN, will be evaluated as a language model, and we will not attempt to simulate human performance such as in situations with garden-path sentences that need to be reanalyzed because of costly initial mis-analysis.

\section{The Differential State Framework and the Delta-RNN}
\label{delta_rnn}
In this section, we will describe the proposed Differential State Framework (DSF) as well as several concrete implementations one can derive from it.

\subsection{General Framework}
\label{framework}
The most general formulation of the architectures that fall under DSF distinguishes two forms of the hidden state. The first is a fast state, which is generally a function of the data at the current time-step and a filtration (or summary function of past states). The second is a slow state, or data-independent state. This concept can be specifically viewed as a composition of two general functions, formally defined as follows:
\begin{align}
\mathbf{h}_{t} &= q_{\Theta}(\mathbf{x}_t,\mathbf{M}_{t-1}) \nonumber \\ &= f_{\psi}[g_{\theta}(\mathbf{x}_t,\mathbf{M}_{t-1}),\mathbf{M}_{t-1}] \label{general_form}
\end{align}
where $\Theta = \{\theta,\psi\}$ are the parameters of the state-machine and $\mathbf{M}_{t-1}$ is the previous latent information the model is conditioned on. In the case of most gated architectures, $\mathbf{M}_{t-1} = \mathbf{h}_{t-1}$, but in some others, as in the SCRN or the LSTM, $\mathbf{M}_{t-1} = \{\mathbf{h}_{t-1},\mathbf{c}_{t-1}\}$\footnote{$\mathbf{c}_t$ refers to the ``cell-state'' as in \parencite{hochreiter1997lstm}.} or could even include information such as de-coupled memory, and in general will be updated as symbols are iteratively processed. We define $g_{\theta}(\cdot)$ to be any, possibly complicated, function that maps the previous hidden state and the currently encountered data point (e.g. a word, subword, or character token) to a real-valued vector of fixed dimensions using parameters $\theta$. $f_{\psi}(\cdot)$, on the other hand, is defined to be the outer function that uses parameters $\psi$ to integrate the fast-state, as calculated by $g_{\theta}(\cdot)$, and the slowly-moving, currently un-transformed state $\mathbf{h}_{t-1}$. In the sub-sections that follow, we will describe simple formulations of these two core functions and, later in Section \ref{rw:recovery}, we will show how currently popular architectures, like the LSTM and various simplifications, are instantiations of this framework. The specific structure of Equation \ref{general_form} was chosen to highlight that we hypothesize the reason behind the success of gated neural architectures is largely because they have been treating the next-step prediction tasks, like language modeling, as an interaction between two functions.  One inner function focuses on integrating observed samples with a current filtration to create a new data-dependent hidden representation (or state ``proposal')' while an outer function focuses on computing the difference, or ``delta'', between the impression of the sub-sequence observed so far (i.e., $\mathbf{h}_{t-1}$) with the newly formed impression. For example, as a sentence is iteratively processed, there might not be much new information (or ``suprisal'') in a token's mapped hidden representation (especially if it is a frequently encountered token), thus requiring less change to the iteratively inferred global representation of the sentence.\footnote{One way to extract a ``sentence representation'' from a temporal neural language model would be to simply to take the last hidden state calculated upon reaching a symbol such as punctuation (e.g., period or exclamation point). This is sometimes referred to as encoding variable-length sentences or paragraphs to a real-valued vector of fixed dimensionality.} However, encountering a new or rare token (especially an unexpected one) might bias the outer function to allow the newly formed hidden impression to more strongly influence the overall impression of the sentence, which will be useful when predicting what token/symbol will come next. In Section \ref{discuss}, we will present a small demonstration using one of the trained word-models to illustrate the intuition just described.

In the sub-sections to follow, we will describe the ways we chose to formulate $g_{\theta}(\cdot)$ and $f_{\psi}(\cdot)$ in the experiments of our paper. The process we followed for developing the concrete implementations of $g_{\theta}(\cdot)$ and $f_{\psi}(\cdot)$ involved starting from the simplest possible form using the fewest (if any) possible parameters to compose each function and testing it in preliminary experiments to verify its usefulness.

It is important to note that Equation \ref{general_form} is still general enough to allow for future design of more clever or efficient functions that might improve the performance and long-term memory capabilities of the framework. More importantly, one might view the parameters $\psi$ that $f_{\psi}(\cdot)$ uses as possibly encapsulating structures that can be used to store explicit memory-vectors, as is the case in stacked-based RNNs \parencite{das1992learning,joulin2015inferring} or linked-list-based RNNs \parencite{joulin2015inferring}.

\subsection{Forms of the Outer Function}
\label{outer_fun}
Keeping $g_{\theta}(\cdot)$ as general as possible, here we will describe several ways one could design $f_{\psi}(\cdot)$, the function meant to decide how new and old hidden representations will be combined at each time step. We will strive to introduce as few additional parameters as necessary and experimental results will confirm the effectiveness of our simple designs.

One form that $f_{\psi}(\cdot)$ could take is a simple weighted summation, as follows:
\begin{align}
\mathbf{h}_{t} &= f_{\psi}[g_{\theta}(\mathbf{x}_t,\mathbf{h}_{t-1}),\mathbf{h}_{t-1}] \nonumber \\ 
&= \Phi(\gamma [g_{\theta}(\mathbf{x}_t,\mathbf{h}_{t-1}) + \beta \mathbf{h}_{t-1}) \label{sum_form}
\end{align}
where $\Phi(\cdot)$ is an element-wise activation applied to the final summation and $\gamma$ and $\beta$ are bias vectors meant to weight the fast and slow states respectively. In Equation \ref{sum_form}, if $\gamma = \beta = 1$, no additional parameters have been introduced making the outer function simply a rigid summation operator followed by a non-linearity. However, one will notice that $\mathbf{h}_{t-1}$ is transmitted across a set of fixed identity connections in addition to being transformed by $g_{\theta}(\cdot)$.

While $\gamma$ and $\beta$ could be chosen to be hyper-parameters and tuned externally (as sort of per-dimension scalar multipliers), it might prove to be more effective to allow the model to learn these coefficients. If we introduce a vector of parameters $\mathbf{r}$, we can choose the fast and slow weights to be $\gamma = (1 -\mathbf{r})$ and $\beta = (\mathbf{r})$, facilitating simple interpolation. Adding these negligibly few additional parameters to compose an interpolation mechanism yields the state-model:
\begin{align}
\mathbf{h}_{t} &= \Phi( (1 - \mathbf{r}) \otimes g_{\theta}(\mathbf{x}_t,\mathbf{h}_{t-1}) + \mathbf{r} \otimes \mathbf{h}_{t-1} ) \mbox{.} \label{gated_form}
\end{align}

Note that we define $\otimes$ to be the Hadamard product. Incorporating this interpolation mechanism can be interpreted as giving the Differential State Framework model a flexible mechanism for mixing various dimensions of its longer-term memory with its more localized memory. Interpolation, especially through a simple gating mechanism, can be an effective way to allow the model to learn how to turn on/off latent dimensions, potentially yielding improved generalization performance, as was empirically shown by \textcite{serban2016multi}.

Beyond fixing $\mathbf{r}$ to some vector of pre-initialized values, there two simple ways to parametrize $\mathbf{r}$:
\begin{align}
\mathbf{r} &=  1 / (1 + exp(-\mathbf{b}_r)) \mbox{, or} \label{gate_param_1}\\
\mathbf{r} &=  1 / (1 + exp(-[W \mathbf{x}_t + \mathbf{b}_r])) \label{gate_param_2}
\end{align}
where both forms only introduce an additional set of learnable bias parameters, however Equation \ref{gate_param_2} allows the data at time step $t$ to interact with the gate and thus takes into account additional information from the input distribution when mixing stable and local states together. Unlike \textcite{serban2016multi}, we constrain the rates to lie in the range $[0,1]$ by using the logistic link function, $\sigma(v) = 1 / (1 + exp(-v))$, which will transform the biases into rates much like the rates of the SCRN. We crucially choose to share $W$ in this particular mechanism for two reasons: 1) we avoid adding yet another matrix of input to hidden parameters and, much to our advantage, reuse the computation of the linear pre-activation term $W \mathbf{x}_t$, and 2) additionally coupling the data pre-activation to the gating mechanism will serve as further regularization of the input-to-hidden parameters (by restricting the amount of learnable parameters, much as in classical autoencoders). Two error signals, $\frac{\partial \mathbf{r}}{\partial W}$ and $\frac{\partial \mathbf{z}_t}{\partial W}$, now take part in the calculation of the partial derivative $\frac{\partial \mathcal{L}(\mathbf{y}_{t},\mathbf{x}_{t+1})}{\partial W}$ ($\mathbf{y}_t$ is the output of the model at $t$).

\begin{figure*}
\fontsize{20}{20}\selectfont 
\centering
 \includegraphics[width=0.985\linewidth]{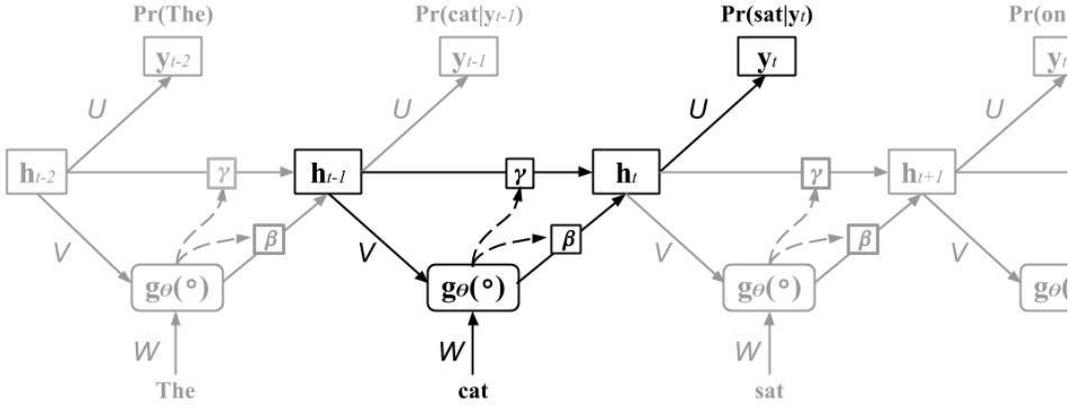}
\caption{The Delta-RNN computation graph, unfolded over time. The learnable gates $\gamma$ and $\beta$ control how much influence the previous state and the currently computed data-dependent state have on computing the model's next hidden state.}
\label{fig:arch}
\end{figure*}

Figure \ref{fig:arch} depicts the architecture using the simple late-integration mechanism.

\subsection{Forms of the Inner Function -- Instantiating the Delta-RNN}
\label{inner_fun}
When a concrete form of the inner function $g_{\theta}(\cdot)$ is chosen, we can fully specify the Differential State Framework. We will also show, in Section \ref{rw:recovery}, how many other commonly-used RNN architectures can, in fact, be treated as special cases of this general framework defined under Equation \ref{general_form}.

Starting from Equation \ref{sum_form}, if we fix $\gamma = 1$ and $\beta = 0$, we can recover the classical Elman RNN, where $g_{\theta}(\mathbf{x}_t,\mathbf{h}_{t-1})$ is a linear combination of the projection of the current data point and the projection of the previous hidden state, followed by a non-linearity $\phi(\cdot)$. However, if we also set $\beta = 1$, we obtain a naive way to compute a delta change of states. Specifically, the simple-RNN's hidden state, where $\Phi(v) = v$ (the identity function), is:
\begin{align}
\mathbf{h}_{t} &= \gamma \otimes g_{\theta}(\mathbf{x}_t,\mathbf{h}_{t-1}) + \beta \otimes \mathbf{h}_{t-1} \nonumber \\
\mathbf{h}_{t} &= 1 \otimes \phi(V \mathbf{h}_{t-1} + W \mathbf{x}_t + \mathbf{b}) + 0 \otimes \mathbf{h}_{t-1} \nonumber \\
\mathbf{h}_{t} &= \phi(V \mathbf{h}_{t-1} + W \mathbf{x}_t + \mathbf{b}) \label{srn}
\end{align}
where $\mathbf{h}_{t}$ is the hidden layer state at time $t$, $\mathbf{x}_{t}$ is the input vector, and $\theta = \{W, V\}$ contains the weight matrices. In contrast, the \emph{simple Delta-RNN}, where instead $\phi(v) = v$, we have:
\begin{align}
\mathbf{h}_{t} = \Phi(V \mathbf{h}_{t-1} + W \mathbf{x}_t + \mathbf{b} + \mathbf{h}_{t-1}) \mbox{.} \label{drnn}
\end{align}
Thus, the state can be implicitly stable, assuming $W$ and $V$ are initialized with small values and $\phi(\cdot)$ allows this by being partially linear. For example we can choose $\phi(\cdot)$ to be the linear rectifier (or initialize the model so to start out in the linear regime of the hyperbolic tangent). In this case, the simple Delta-RNN does not need to learn anything to maintain the state constant over time. 

Preliminary experimentation with this simple form (Equation \ref{drnn}) often yielded unsatisfactory performance. This further motivated the development of the simple interpolation mechanism presented in Equation \ref{gated_form}. However, depending on how one chooses the non-linearities, $\phi(\cdot)$ and $\Phi(\cdot)$, one can create different types of interpolation. Using an Elman RNN for $g_{\theta}(\mathbf{x}_t,\mathbf{h}_{t-1})$ as in Equation \ref{srn}, substituting into Equation \ref{gated_form} can create what we propose as the ``late-integration'' state model:
\begin{align}
\mathbf{z}_{t} &= g_{\theta}(\mathbf{x}_t,\mathbf{h}_{t-1}) \nonumber \\
&= \phi(V \mathbf{h}_{t-1} + W \mathbf{x}_t + \mathbf{b}) \mbox{,  and,} \\
\mathbf{h}_{t} &=  \Phi( (1 - \mathbf{r}) \otimes \mathbf{z}_{t} + \mathbf{r} \otimes \mathbf{h}_{t-1} ) \mbox{.} \label{late_integration}
\end{align}
where $\Phi(\cdot)$ could be any choice of activation function, including the identity function. This form of interpolation allows for a more direct error propagation pathway since gradient information, once transmitted through the interpolation gate, has two pathways: through the non-linearity of the local state (through $g_{\theta}(\mathbf{x}_t,\mathbf{h}_{t-1})$) and the pathway composed of implicit identity connections.\footnote{Late-integration might remind the reader of the phrase ``late fusion'', as in the context of \textcite{wang2015larger}.  However, \citeauthor{wang2015larger} was focused on merging the information from an external bag-of-words context vector with the standard cell state of the LSTM.}

When using a simple Elman RNN, we have essentially described a first-order Delta-RNN.  However, historically, second-order recurrent neural architectures have been shown to be powerful models in tasks such as grammatical inference \parencite{giles1991second} and noisy time-series prediction \parencite{giles2001noisy} as well as incredibly useful in rule-extraction when treated as finite-state automata \parencite{giles1992learning,goudreau1994first}. Very recently, \textcite{wu_multiplicative_2016} showed that the gating effect between the state-driven component and data-driven components of a layer's pre-activations facilitated better propagation of gradient signals as opposed to the usual linear combination. A second-order version of $g_{\theta}(\mathbf{x}_t,\mathbf{h}_{t-1})$ would be highly desirable, not only because it further mitigates the vanishing gradient problem that plagues back-propagation through time (used in calculating parameter gradients of neural architectures), but because the form introduces negligibly few additional parameters. We do note that the second-order form we use, like in \textcite{wu_multiplicative_2016}, is a rank-1 matrix approximation of the actual tensor used in \textcite{giles1992learning,goudreau1994first}.

We can take the late-integration model, Equation \ref{late_integration}, and replace, similar to \textcite{giles1991second}, $\mathbf{z}_t$ with:
\begin{align}
\mathbf{z}_{t} &= \phi(V \mathbf{h}_{t-1} \otimes W \mathbf{x}_t + \mathbf{b}) \label{second_order}
\end{align}
or a more general form \parencite{wu_multiplicative_2016}:
\begin{align}
\mathbf{d}^1_t &= \alpha \otimes V_d \mathbf{h}_{t-1} \otimes W \mathbf{x}_t \nonumber \\
\mathbf{d}^2_t &= \beta_1 \otimes V_d \mathbf{h}_{t-1} + \beta_2 \otimes W \mathbf{x}_t \nonumber \\
\mathbf{z}_{t} &= \phi(\mathbf{d}^1_t + \mathbf{d}^2_t + \mathbf{b})\label{general_second_order},
\end{align}
where we note that $\mathbf{z}_t$ can be a function of any arbitrary incoming set of information signals that are gated by the last known state. The \emph{Delta-RNN} will ultimately combine this data-driven signal $\mathbf{z}_t$ with its slow-moving state. More importantly, observe that even in the most general form (Equation \ref{general_second_order}), only a few further bias vector parameters, $\alpha$, $\beta_1$, and $\beta_2$ are required. 

Assuming a single hidden layer language model, with $H$ hidden units and $V$ input units (where $V$ corresponds to the  cardinality of the symbol dictionary), \textbf{a full late-integration Delta-RNN that employs a second-order $g_{\theta}(\mathbf{x}_t,\mathbf{h}_{t-1})$ (Equation \ref{general_second_order}), has only $((H * H) + 2(H * V) + 5H + V)$ parameters\footnote{$5H$ counts the hidden bias, the full interpolation mechanism $\mathbf{r}_t$ (Equation \ref{gate_param_2}), and the second-order biases, $\{\alpha,\beta_1,\beta_2\}$.}, which is only slightly larger than a classical RNN with only $((H * H) + 2(H * V) + H + V)$ parameters.} This stands in stark contrast to the sheer number of parameters required to train commonly-used complex architectures such as the LSTM (with peephole connections), with $(4(H * H) + 8(H * V) + 4H + V)$ parameters, and the GRU, with $(3(H * H) + 4(H * V) + 3H + V)$ parameters. 

\subsection{Regularizing the Delta-RNN}
\label{reg}
Regularization is often important when training large, over-parametrized models. To control for overfitting, approaches range from structural modifications to impositions of priors over parameters \citep{neal2012bayesian}. Commonly employed modern approaches include drop-out \citep{srivastava2014dropout} and variations \citep{gal2016theoretically} or mechanisms to control for internal covariate drift, such as Batch Normalization \citep{ioffe2015batch} for large feedforward architectures. In this paper, we investigate the effect that drop-out will have on the Delta-RNN's performance.\footnote{In preliminary experiments, we also investigated incorporating layer normalization \citep{ba2016layer} into the Delta-RNN architecture, the details of which may be found in the Appendix.  We did not find observe noticeable gains using layer normalization over drop-out, and thus only report the results of drop-out in this paper.}

To introduce simple (non-recurrent) drop-out to the framework, our preliminary experiments uncovered that drop-out was most effective when applied to the inner function $g(\cdot)$ as opposed to the outer function's computed delta-state. For the full Delta-RNN, under drop-out probability $p_{drop}$, this would lead to the following modification:
\begin{align}
\mathbf{h}_{t} &= \Phi( (1 - \mathbf{r}) \otimes DROP(g_{\theta}(\mathbf{x}_t,\mathbf{h}_{t-1}), p_{drop}) + \mathbf{r} \otimes \mathbf{h}_{t-1} ) \mbox{.} \label{dropout_form}
\end{align}
$DROP(\mathbf{x}, p_{drop}) = \mathbf{x}  \otimes (\sim\mathbf{B}(1, p_{drop}))$ is the drop-out operator which masks its input argument with a binary vector sampled from $H$ independent Bernoulli distributions. 

\subsection{Learning under the Delta-RNN}
\label{learning}
Let $w_1, \dots, w_N$ be a variable-length sequence of $N$ symbols (such as words that would compose a sentence). In general, the distribution over the variables follows the graphical model:
\begin{align}
P_{\theta}(w_1, \dots, w_T) = \prod_{t=1}^T P_{\Theta}(w_t | w_{<t}),
\end{align}
where $\Theta = \{ \psi,\theta \} = \{V,W,R,\mathbf{b},\mathbf{b}_r,\alpha,\beta_1,\beta_2\}$ are the model parameters (of a full Delta-RNN).

No matter how the hidden state $\mathbf{h}_t$ is calculated, in this paper, it will ultimately be fed into a maximum-entropy classifier\footnote{Note that the bias term has been omitted for clarity.} defined as:
\begin{align}
P(w,\mathbf{h}_t) = P_{\Theta}(w|\mathbf{h}_t) =  \frac{\exp{( w^{\text{T}} R \mathbf{h}_t)}}{\sum_{w'} \exp{(w^{\text{T}} R \mathbf{h}_t)}}, \label{max_ent_1}
\end{align}
To learn parameters for any of our  models, we optimize with respect to the sequence negative log likelihood:
\begin{align}
\mathcal{L} = -\sum^{N}_{i=1} \sum^{T}_{t=1} \log P_{\Theta}(w_t | \mathbf{h}),
\end{align}
Model parameters, $\Theta = \{\theta, \psi\}$, of the Delta-RNN are learned under an empirical risk minimization framework.  We employ back-propagation of errors (or rather, reverse-mode automatic differentiation with respect to this negative log likelihood objective function) to calculate gradients and update the parameters using the method of steepest gradient descent. For all experiments conducted in this paper, we found that the ADAM adaptive learning rate scheme \parencite{kingma2014adam} (followed by a Polyak average \parencite{polyak1992acceleration} for the subword experiments) yielded the most consistent and near-optimal performance. We therefore use this set-up for optimization of parameters for all models (including baselines), unless otherwise mentioned. For all experiments, we unroll computation graphs $T$ steps in time (where $T$ varies across experiments/tasks), and, in order to approximate full back-propagation through time, we carry over the last hidden from the previous mini-batch (within a full sequence). More importantly, we found that by furthermore using the derivative of the loss with respect to the last hidden state, we can improve the approximation and thus perform one step of iterative inference \footnote{We searched the step-size $\lambda$ over the values $\{0.05, 0.1, 0.15\}$ for all experiments in this paper.} to update the last hidden state carried over. We ultimately used this proposed improved approximation for the sub-word models (since in those experiments we could directly train all baseline and proposed models in a controlled, identical fashion to ensure fair comparison).

For all Delta-RNNs experimented with in this paper, the output activation of the inner function $g(\cdot)$ was chosen to be the hyperbolic tangent. The output activation of the outer function $f(\cdot)$ was set to be the identity for the word and character benchmark experiments and the hyperbolic tangent for the subword experiments (these decisions were made based on preliminary experimentation on sub-sets of the final training data). The exact configuration of the implementation we used in this paper involved using the late-integration form, either the un-regularized (Equation \ref{late_integration}) or the drop-out regularized (Equation \ref{dropout_form}) variant, for the outer function and Equation \ref{general_second_order}

We compare our proposed models against a wide variety of un-regularized baselines, as well several state-of-the-art regularized baselines for the benchmark experiments.
These baselines include the LSTM, GRU, and SCRN as well as computationally more efficient formulations of each, such as the MGU. The goal is to see if our proposed Delta-RNN is a suitable replacement for complex gated architectures and can capture longer term patterns in sequential text data.

\section{Related Work: Recovering Previous Models}
\label{rw:recovery}
A contribution of this work is that our general framework, presented in Section \ref{framework}, offers a way to unify previous proposals for gated neural architectures (especially for use in next-step prediction tasks like language modeling) and explore directions of improvement. Since we will ultimately compare our proposed Delta-RNN of Section \ref{inner_fun} to these architectures, we will next present how to derive several key architectures from our general form, such as the Gated Recurrent Unit and the Long Short Term Memory. More importantly, we will introduce them in the same notation / design as the Delta-RNN and highlight the differences between previous work and our own through the lens of $f_{\psi}(\cdot)$ and $g_{\theta}(\mathbf{x}_t,\mathbf{M}_{t-1})$.

Simple models, largely based on the original Elman RNN \parencite{elman1990finding}, have often been shown to perform quite well in language modeling tasks \parencite{mikolov2010recurrent,mikolov2011extensions}. The Structurally Constrained Recurrent Network (SCRN,  \cite{mikolov2014learning}), an important predecessor and inspiration for this work, showed that one fruitful path to learning longer-term dependencies was to impose a hard constraint on how quickly the values of hidden units could change, yielding more ``stable'' long-term memory. The SCRN itself is very similar to a combination of the RNN architectures of \parencite{jordan1986attractor,mozer1993neural}. The key element of its design is the constraint that part of recurrent weight matrix must stay close to the identity, a constraint that is also satisfied by the Delta-RNN. These identity connections (and corresponding context units that use them) allow for improved information travel over many time-steps and can even be viewed as an exponential trace memory \parencite{mozer1993neural}. Residual Networks, though feed-forward in nature, also share a similar motivation \parencite{he2016identity}. Unlike the SCRN, the proposed Delta-RNN does not require a separation of the slow and fast moving units, but instead models this slower time-scale through implicitly stable states.

The Long Short Term Memory (LSTM, \cite{hochreiter1997long}) is arguably the currently most popular and often-used gated neural architecture, especially in the domain of Natural Language Processing. Starting from our general form, Equation \ref{general_form}, we can see how the LSTM can be deconstructed, where setting $\mathbf{c}_t = g_{\theta}(\mathbf{x}_t,\mathbf{M}_{t-1})$, yields:
\begin{align}
\mathbf{h}_{t} &= f_{\psi}[g_{\theta}(\mathbf{x}_t,\mathbf{M}_{t-1}),\mathbf{M}_{t-1}] \nonumber \\
\mathbf{h}_{t} &= \mathbf{r}_t \otimes \Phi(\mathbf{c}_t)  \mbox{, where,} \\ 
\mathbf{r}_{t} &= \sigma(W_r \mathbf{x}_t + V_r \mathbf{h}_{t-1} + U_r \mathbf{c}_t + \mathbf{b}_r) \label{lstm_outer}
\end{align}
where $\mathbf{M}_{t-1} = \{\mathbf{h}_{t-1}, \mathbf{c}_{t-1}\}$, noting that $\mathbf{c}_{t-1}$ is the cell-state designed to act as the constant error carousal in mitigating the problem of vanishing gradients when using back-propagation through time. A great deal of recent work has attempted to improve the training of the LSTM, often by increasing its complexity, such as through the introduction of so-called ``peephole connections'' \parencite{gers2000recurrent}. To compute $\mathbf{c}_t = g_{\theta}(\mathbf{x}_t,\mathbf{M}_{t-1})$, using peephole connections, we use the following set of equations:
\begin{align*} 
\mathbf{c}_t &= \mathbf{f}_t \otimes \mathbf{c}_{t-1} + \mathbf{i}_t \otimes \mathbf{z}_t \mbox{, where,} \\
\mathbf{z}_t &= \Phi(W_z \mathbf{x}_t + V_z \mathbf{h}_{t-1} + \mathbf{b}_z) \mbox{,} \\
\mathbf{i}_t &= \sigma(W_i \mathbf{x}_t + V_i \mathbf{h}_{t-1} + U_i \mathbf{c}_{t-1} + \mathbf{b}_i) \mbox{,} \\
\mathbf{f}_t &= \sigma(W_f \mathbf{x}_t + V_f \mathbf{h}_{t-1} + U_f \mathbf{c}_{t-1} + \mathbf{b}_f) \mbox{.}
\end{align*}
The Gated Recurrent Unit (GRU, \cite{chung2014empirical,chung2015gated}) can be viewed as one of the more successful attempts to simplify the LSTM. We see that $f_{\psi}(\cdot)$ and $g_{\theta}(\cdot)$ are still quite complex, requiring many intermediate computations to reach an output. In the case of the outer mixing function, $f_{\psi}(\cdot)$, we see that:
\begin{align}
\mathbf{h}_{t} &= \Phi(\gamma [g_{\theta}(\mathbf{x}_t,\mathbf{h}_{t-1}) + \beta \mathbf{h}_{t-1}) \nonumber \\
\gamma &= \mathbf{r}_t \mbox{ and } \beta =  (1 - \mathbf{r}_t) \mbox{, where,} \\
\mathbf{r}_t &= \sigma(V_{r} \mathbf{h}_{t-1} + W_{r} \mathbf{x}_t + \mathbf{b}_{r}) \label{gru_outer}
\end{align}
noting that the state gate $\mathbf{r}_t$ is also a function of the RNN's previous hidden state and introduces parameters specialized for $\mathbf{r}$. In contrast, the Delta-RNN does not use an extra set of input-to-hidden weights, and more directly, the pre-activation of the input projection can be reused for the interpolation gate. The inner function of the GRU, $g_{\theta}(\mathbf{x}_t,\mathbf{h}_{t-1})$, is  defined as:
\begin{align*}
g_{\theta}(\mathbf{x}_t,\mathbf{h}_{t-1}) &= \phi(V_h (\mathbf{q}_t \otimes \mathbf{h}_{t-1}) + W_h \mathbf{x}_t + \mathbf{b}_h) \\
\mathbf{q}_t &= \sigma(V_q \mathbf{h}_{t-1} + W_q \mathbf{x}_t + \mathbf{b}_q) 
\end{align*}
where $\phi()$ is generally set to be the hyperbolic tangent activation function. A mutated architecture (MUT, \cite{zaremba2015empirical}) was an attempt to simplify the GRU somewhat, as, much like the Delta-RNN, its interpolation mechanism is not a function of the previous hidden state but is still largely as parameter-heavy as the GRU, only shedding a single extra parameter matrix, especially since its interpolation mechanism retains a specialized parameter matrix to transform the data. The Delta-RNN, on the other hand, shares this with its primary calculation of the data's pre-activation values. The Minimally Gated Unit (MGU,  \cite{zhou2016minimal}) is yet a further attempt to reduce the complexity of the GRU by merging its reset and update gates into a single \emph{forget} gate, essentially using the same outer function under the GRU defined in Equation \ref{gru_outer}, but simplifying the inner function $g_{\theta}(\mathbf{x}_t,\mathbf{h}_{t-1})$ to be quite close to the Elman-RNN but conditioned on the forget gate as follows:
\begin{align*}
g_{\theta}(\mathbf{x}_t,\mathbf{h}_{t-1}) &= \phi(V_h (\mathbf{r}_t \otimes \mathbf{h}_{t-1}) + W_h \mathbf{x}_t + \mathbf{b}_h) \mbox{.}
\end{align*}
While the MGU certainly does reduce the number of parameters
, viewing it from the perspective of our general Delta-RNN framework, one can see that it still largely uses a $g_{\theta}(\mathbf{x}_t,\mathbf{h}_{t-1})$ that is rather limited (only the capabilities of the Elman-RNN). 
The most effective version of our Delta-RNN emerged from the insight that a more powerful $g_{\theta}(\mathbf{x}_t,\mathbf{h}_{t-1})$ could be obtained by (approximately) increasing its order, which requires a few more bias parameters, and nesting it within a non-linear interpolation mechanism that will compute the delta-states. Our framework is general enough to also allow designers to incorporate functions that augment the general state-engine with an external memory to create architectures that can exploit the strengths of models with decoupled memory architectures \parencite{weston_memory_2014,sukhbaatar_end_to_end_2015,graves2016hybrid} or data-structures that serve as memory \parencite{sun1998neural,joulin2015inferring}.

A final related, but important, strand of work uses depth (i.e., number of processing layers) to directly model various time-scales, as emulated in models such as the hierarchical/multi-resolutional recurrent neural network (HM-RNN) \parencite{chung2016hierarchical}. Since the Delta-RNN is designed to allow its interpolation gate $\mathbf{r}$ to be driven by the data, it is possible that the model might already be learning how to make use of boundary information (word boundaries at the character/sub-word level, sentence boundaries as marked by punctuation at the word-level). The HM-RNN, however, more directly attacks this problem  by modifying an LSTM to learn how to manipulate its states when certain types of symbols are encountered. (This is different from models like the Clockwork RNN that require explicit boundary information \parencite{koutnik2014clockwork}.)  One way to take advantage of the ideas behind the HM-RNN would be to manipulate the Differential State Framework to incorporate the explicit modeling of time-scales through layer depth (each layer is responsible for modeling a different time-scale). Furthermore, it would be worth investigating how the HM-RNN's performance would change when built from modifying a Delta-RNN instead of an LSTM. 

\section{Experimental Results}
\label{experiments}
Language modeling is an incredibly important next-step prediction task, with applications in downstream applications in speech recognition, parsing, and information retrieval. As such, we will focus this paper on experiments on this task domain to gauge the efficacy of our Delta-RNN framework, noting that the Delta-RNN framework might prove useful in, for instance, machine translation \parencite{bahdanau2014neural} or light chunking \parencite{turian2009quadratic}. Beyond improving language modeling performance, the sentence (and document) representations iteratively inferred by our architectures might also prove useful in composing higher-level representations of text corpora, a subject we will investigate in future work.

\subsection{Datasets}

\subsubsection{The Penn Treebank Corpus}
\label{corpus:ptb}
The Penn Treebank corpus \parencite{marcus1993building} is often used to benchmark both word and character-level models via perplexity or bits-per-character, and thus we start here.\footnote{To be directly comparable with previously reported results, we make use of the specific pre-processed train/valid/test splits found at http://www.fit.vutbr.cz/~imikolov/rnnlm/.} The corpus contains 42,068 sentences (971,657 tokens, average token-length of about 4.727 characters) of varying length (the range is from 3 to 84 tokens, at the word-level).

\subsubsection{The IMDB Corpus}
\label{imdb}
The large sentiment analysis corpus \parencite{maas_2011_ACL_HLT2011} is often used to benchmark algorithms for predicting the positive or negative tonality of documents.  However, we opt to use this large corpus (training consists of 149,714 documents, 1,875,523 sentences, 40,765,697 tokens, average token-length is about 3.4291415 characters) to evaluate our proposed Delta-RNN as a (subword) language model. The IMDB data-set serves as a case when the context extends beyond the sentence-level in the form of actual documents.

\begin{table*}[!t]
\renewcommand{\arraystretch}{0.485}
\caption{Test-set results on the Penn Treebank word-level and character-level language modeling tasks. Note that ``impl.'' means implementation.}
\centering
\label{results:ptb_word}
\centering
\begin{tabular}{ll}

\textbf{Penn Treebank: Word Models}&\textbf{PPL} \\
\hline
\emph{N-Gram } \parencite{mikolov2014learning} & $141$ \tabularnewline
\emph{NNLM } \parencite{mikolov_thesis} & $140.2$ \tabularnewline
\emph{N-Gram+cache } \parencite{mikolov2014learning} & $125$ \tabularnewline
\emph{RNN } \parencite{gulcehre2016noisy} & $129$ \tabularnewline
\emph{RNN } \parencite{mikolov_thesis} & $124.7$ \tabularnewline
\emph{LSTM } \parencite{mikolov2014learning} & $115$ \tabularnewline
\emph{SCRN } \parencite{mikolov2014learning} & $115$ \tabularnewline
\emph{LSTM } \parencite{sundermeyer2016} & $107$\tabularnewline
\emph{MI-RNN} (\cite{wu_multiplicative_2016}, our impl.) & $109.2$ \tabularnewline
\emph{Delta-RNN} (present work) & $100.324$\tabularnewline 
\emph{Delta-RNN, dynamic \#1} (present work) & $93.296$\tabularnewline
\emph{Delta-RNN, dynamic \#2} (present work) & $90.301$\tabularnewline

\hline

&  \tabularnewline 
\emph{LSTM-recurrent drop } \parencite{krueger2016zoneout} & $87.0$ \tabularnewline
\emph{NR-dropout } \parencite{zaremba2014recurrent} & $78.4$ \tabularnewline
\emph{V-dropout } \parencite{gal2016theoretically} & $73.4$ \tabularnewline
\emph{Delta-RNN-drop, static} (present work) & $84.088$\tabularnewline
\emph{Delta-RNN-drop, dynamic \#1} (present work) & $79.527$\tabularnewline
\emph{Delta-RNN-drop, dynamic \#2} (present work) & $78.029$\tabularnewline
\end{tabular}

\label{results:ptb_char}
\centering
\vskip1em
\begin{tabular}{ll}

\textbf{Penn Treebank: Character Models}&\textbf{BPC}
\tabularnewline
\hline
\emph{N-discount N-gram} \parencite{mikolov2012subword} & $1.48$ \tabularnewline
\emph{RNN+stabilization} (Krueger et al., 2016) & $1.48$ \tabularnewline 
\emph{linear MI-RNN} \parencite{wu_multiplicative_2016} & $1.48$ \tabularnewline
\emph{Clockwork RNN} \parencite{koutnik2014clockwork} & $1.46$ \tabularnewline
\emph{RNN } \parencite{mikolov2012subword} & $1.42$ \tabularnewline
\emph{GRU } \parencite{jernite2016variable} & $1.42$ \tabularnewline
\emph{HF-MRNN } \parencite{mikolov2012subword} & $1.41$ \tabularnewline
\emph{MI-RNN } \parencite{wu_multiplicative_2016} & $1.39$ \tabularnewline
\emph{Max-Ent N-gram} \parencite{mikolov2012subword} & $1.37$ \tabularnewline
\emph{LSTM } \parencite{krueger2016zoneout} & $1.356$ \tabularnewline
\emph{Delta-RNN} (present work) & $1.347$ \tabularnewline 
\emph{Delta-RNN, dynamic \#1} (present work) & $1.331$\tabularnewline
\emph{Delta-RNN, dynamic \#2} (present work) & $1.326$\tabularnewline

\hline

&  \tabularnewline 
\emph{LSTM-norm stabilizer } \parencite{krueger2016zoneout} & $1.352$ \tabularnewline
\emph{LSTM-weight noise } \parencite{krueger2016zoneout} & $1.344$ \tabularnewline
\emph{LSTM-stochastic depth } \parencite{krueger2016zoneout} & $1.343$ \tabularnewline
\emph{LSTM-recurrent drop } \parencite{krueger2016zoneout} & $1.286$ \tabularnewline
\emph{RBN } \parencite{cooijmans2016recurrent} & $1.32$ \tabularnewline
\emph{LSTM-zone out } \parencite{krueger2016zoneout} & $1.252$ \tabularnewline
\emph{H-LSTM + LN  } \parencite{ha2016hypernetworks} & $1.25$ \tabularnewline
\emph{TARDIS} \parencite{gulcehre2017memory} & $1.25$ \tabularnewline
\emph{3-HM-LSTM + LN } \parencite{chung2016hierarchical} & $1.24$ \tabularnewline
\emph{Delta-RNN-drop, static} (present work) & $1.251$ \tabularnewline
\emph{Delta-RNN-drop, dynamic \#1} (present work) & $1.247$\tabularnewline
\emph{Delta-RNN-drop, dynamic \#2} (present work) & $1.245$\tabularnewline
\end{tabular}

\end{table*}

\subsection{Word \& Character-Level Benchmark}
\label{benchmarks}
The first set of experiments allow us to examine our proposed Delta-RNN models against reported state-of-the-art models. These reported measures have been on traditional word and character-level language modeling tasks--we measure the per-symbol perplexity of models. For the word-level models, we calculate the per-word perplexity (PPL) using the measure $PPL = \exp \big [-(1/N) \sum^{N}_{i=1} \sum^{T}_{t=1} \log P_{\Theta}(w_t | \mathbf{h}) \big ]$. For the character-level models, we report the standard bits-per-character (BPC), which can be calculated from the log likelihood using the formula: $BPC = -1/(N \log(2)) \sum^{N}_{i=1} \sum^{T}_{t=1} \log P_{\Theta}(w_t | \mathbf{h})$.

Over 100 epochs, word-level models with mini-batches of 64 (padded) sequences. (Early stopping with a lookahead of 10 was used.) Gradients were clipped using a simple magnitude-based scheme \parencite{pascanu2013difficulty}, with the magnitude threshold set to 5.  A simple grid-search was performed to tune the learning rate, $\lambda = \{0.002, 0.001, 0.0005, 0.0002\}$, as well as the size of the hidden layer $H = \{500,1000,1500\}$. Parameters (non-biases) were initialized from zero-mean Gaussian distributions with variance tuned, $\sigma = \{0.1, 0.01, 0.005, 0.001\}$\footnote{We also experimented with other initializations, most notably the identity matrix for the recurrent weight parameters as in \parencite{le2015simple}. We found that this initialization often worsened performance. For the activation functions of the first-order models, we experimented with the linear rectifier, the parametrized linear rectifier, and even our own proposed parametrized smoothened linear rectifier, but found such activations lead to less-than-satisfactory results. The results of this inquiry is documented in the code that will accompany the paper.}. The character-level models, on the other hand, were updated using mini-batches of 64 samples over 100 epochs. (Early stopping with a lookahead of 10 was used.) The parameter initializations and grid-search for the learning rate and hidden layer size were the same as for the word models, with the exception of the hidden layer size, which was searched over $H = \{500,1000,1500,2000\}$\footnote{Note that $H = 2000$ would yield nearly 4 million parameters, which was our upper bound on total number of parameters allowed for experiments in order to be commensurable with the work of \textcite{wu_multiplicative_2016}, which actually used $H = 2048$ for all Penn Treebank models. }. 

A simple learning rate decay schedule was employed: if the validation loss did not decrease after a single epoch, the learning rate was halved (unless a lower bound on the value had been reached). When drop-out was applied to the Delta-RNN (\emph{Delta-RNN-drop}, we set the probability of dropping a unit to $p_{drop} = 0.15$ for the character-level models and $p_{drop} = 0.5$ for the word level models. We present the results for the un-regularized and regularized versions of the models. For all of the Delta-RNNs, we furthermore experiment with two variations of dynamic evaluation, which facilitates fair comparison to compression algorithms, inspired by the improvements observed in \parencite{mikolov_thesis}. \emph{Delta-RNN-drop, dynamic \#1} refers to simply updating the model sample-by-sample after each evaluation, where in this case, we update parameters using simple stochastic gradient descent \parencite{mikolov_thesis}, with a step-size $\lambda = 0.005$. We develop a second variation of dynamic evaluation, \emph{Delta-RNN-drop, dynamic \#2}, where we allow the model to first iterate (and update) once over the validation set and then finally the test-set, completely allowing the model to ``compress'' the Penn Treebank corpus. These two schemes are used for both the word and character-level benchmarks. It is important to stress the BPC and PPL measures reported for the dynamic models follow a strict ``test-then-train'' online paradigm, meaning that each next-step prediction is made before updating model parameters.

The standard vocabulary for the word-level models contains 10K unique words (including an unknown token for out-of-vocabulary symbols and an end-of-sequence token)\footnote{We use a special ``null'' token (or zero-vector) to mark the start of a sequence.} and the standard vocabulary for the character-level models includes 49 unique characters (including a symbol for spaces). Results for the word-level models are reported in Table \ref{results:ptb_word} and results for the character-level models are reported in Table \ref{results:ptb_char}.

\subsection{Sub-word Language Modeling}
\label{sw_modeling}

We chose to measure the negative log likelihood of the various architectures in the task of \emph{subword modeling}. Subwords are particularly appealing not only in that the input distribution is of lower dimensionality but, as evidenced by the positive results of \textcite{mikolov2012subword}, sub-word/character hybrid language models improve over the performance of pure character-level models. Sub-word models also enjoy the advantage held by character-level models when it comes to handling out-of-vocabulary words, avoiding the need for an ``unknown'' token. 
Research in psycholinguistics has long suggested that even human infants are sensitive to word boundaries at an early stage (e.g., \cite{aslin1998infants}), and that morphologically complex words enjoy dedicated processing mechanisms \parencite{baayen2006morph}. Subword-level language models may approximate such an architecture. Consistency in subword formation is critical in order to obtain meaningful results \parencite{mikolov2012subword}. Thus, we design our sub-word algorithm to partition a word according to the following scheme:

\begin{enumerate}[itemsep=0.0cm]
  \item Split on vowels (using a predefined list)
  \item Link/merge each vowel with a consonant to the immediate right if applicable
  \item Merge straggling single characters to subwords on the immediate right unless a subword of shorter character length is to the left.
\end{enumerate}

This simple partitioning scheme was designed to ensure that no subword was shorter than two characters in length. Future work will entail designing a more realistic subword partitioning algorithm. Subwords below a certain frequency were discarded, and combined with 26 single characters to create the final dictionary. For Penn Treebank, this yields a vocabulary of 2405 symbols was created (2,378 subwords + 26 characters + 1 end-token). For the IMDB corpus, after replacing all emoticons and special non-word symbols with special tokens, we obtain a dictionary of 1926 symbols (1899 subwords + 26 single characters + 1 end-token). Results for all sub-word models are reported in Table~\ref{subword_results}.

\begin{table*}[!t]
\renewcommand{\arraystretch}{.5}
\caption{Test-set negative log likelihoods while holding number of parameters approximately constant.  Subword modeling tasks on Penn Treebank and IMDB.}
\label{subword_results}
\centering
\begin{floatrow}

\begin{tabular}{lll}
\multicolumn{1}{l}{\begin{tabular}[x]{@{}c@{}}\textbf{PTB-SW}\\\end{tabular}}&\multicolumn{1}{c}{\begin{tabular}[x]{@{}c@{}}\textbf{Performance }\\\end{tabular}}&\multicolumn{1}{c}{\begin{tabular}[x]{@{}c@{}}\textbf{}\\\end{tabular}}\tabularnewline
\hline 
\textit{} & \# Params & NLL \tabularnewline
\textit{RNN} & $1,272,464$ & $1.8939$\tabularnewline
\textit{SCRN} & $1,268,604$ & $1.8420$\tabularnewline 
\textit{MGU} & $1,278,692$ & $1.8694$ \tabularnewline
\textit{MI-RNN} & $1,267,904$ & $1.8441$ \tabularnewline
\textit{GRU} & $1,272,404$ & $1.8251$ \tabularnewline
\textit{LSTM} & $1,274,804$ & $1.8412$\tabularnewline 
\textit{Delta-RNN} & $1,268,154$ & $1.8260$ \tabularnewline

\end{tabular}
\centering
\begin{tabular}{lll}
\multicolumn{1}{l}{\begin{tabular}[x]{@{}c@{}}\textbf{IMDB-SW}\\\end{tabular}}&\multicolumn{1}{c}{\begin{tabular}[x]{@{}c@{}}\textbf{Performance }\\\end{tabular}}&\multicolumn{1}{c}{\begin{tabular}[x]{@{}c@{}}\textbf{}\\\end{tabular}}\tabularnewline
\hline
\textit{} & \# Params & NLL \tabularnewline
\textit{RNN} & $499,176$ & $2.1691$\tabularnewline
\textit{SCRN} & $496,196$ & $2.2370$\tabularnewline
\textit{MGU} & $495,444$ & $2.1312$\tabularnewline
\textit{MI-RNN} & $495,446$ & $2.1741$\tabularnewline
\textit{GRU} & $499,374$ & $2.1551$\tabularnewline
\textit{LSTM} & $503,664$ & $2.2080$\tabularnewline
\textit{Delta-RNN} & $495,570$ & $2.1333$\tabularnewline

\end{tabular}
\end{floatrow}
\end{table*}

Specifically, we test our implementations of the LSTM \footnote{We experimented with initializing the forget gate biases of all LSTMs with values searched over $\{1,2,3\}$ since previous work has shown this can improve model performance.} (with peephole connections as described in \cite{graves2013generating}), the GRU, the MGU, the SCRN, as well as a classical Elman network, of both 1st and 2nd-order \parencite{giles1991second,wu_multiplicative_2016}.\footnote{We will publicly release code to build and train the architectures in this paper upon publication.} Subword models were trained in a similar fashion as the character-level models, updated (every 50 steps) using mini-batches of 20 samples but over 30 epochs. Learning rates were tuned in the same fashion as the word-level models, and the same parameter initialization schemes were explored. The notable difference between this experiment and the previous ones is that we fix the number of parameters for each model to be equivalent to that of an LSTM with 100 hidden units for PTB and 50 hiddens units for IMDB. This ensures a controlled, fair comparison across models and allows us to evaluate if the Delta-RNN can learn similarly to models with more complicated processing elements (an LSTM cell versus a GRU cell versus a Delta-RNN unit). Furthermore, this allows us to measure parameter efficiency, where we can focus on the value of actual specific cell-types (for example, allowing us to compare the value of a much more complex LSTM memory unit versus a simple Delta-RNN cell) when the number of parameters is held roughly constant.  We are currently running larger versions of the models depicted in the table above to determine if the results hold at scale.

\section{Discussion} 
\label{discuss}
With respect to the word and character-level benchmarks, we see that the Delta-RNN outperforms all previous, un-regularized models, and performs comparably to regularized state-of-the-art. As documented in Table \ref{subword_results}, we further trained a second-order, word-level RNN (MI-RNN) to complete the comparison, and remark that the second-order connections appear to be quite useful in general, outperforming the SCRN and coming close to that of the LSTM. This extends the results of \textcite{wu_multiplicative_2016} to the word-level. However, the Delta-RNN, which also makes use of second-order units within its inner function, ultimately offers the best performance and performs better than the LSTM in all experiments. In both Penn Treebank and IMDB subword language modeling experiments, the Delta-RNN is competitive with complex architectures such as the GRU and the MGU. In both cases, the Delta-RNN nearly reaches the same performance as the best performing baseline model in either data-set (i.e., it nearly reaches the same performance as the GRU on Penn Treebank and the MGU on IMDB). Surprisingly, on IMDB, a simple Elman network is quite performant, even outperforming the MI-RNN. We argue that this might be the result of constraining all neural architectures to only a small number of parameters for such a large data-set, a constraint we intend to relax in future work.

The Delta-RNN is far more efficient than a complex LSTM and certainly a memory-augmented network like TARDIS \parencite{gulcehre2017memory}. Moreover, it appears to learn how to make appropriate use of its interpolation mechanism to decide how and when to update its hidden state in the presence of new data.\footnote{At greater computational cost, a somewhat lower perplexity for an LSTM may be attainable, such as the perplexity of 107 reported by \textcite{sundermeyer2016} (see Table~\ref{results:ptb_word}).  However, this requires many more training epochs and precludes batch training.} Given our derivations in Section \ref{rw:recovery}, one could argue that nearly all previously proposed gated neural architectures are essentially trying do the same thing under the Differential State Framework. The key advantage offered by the Delta-RNN is that this functionality is offered directly and cheaply (in terms of required parameters).

It is important to contrast these (un-regularized) results with those that use some form of regularization. \textcite{zaremba2014recurrent} reported that a single LSTM (for word-level Penn Treebank) can reach a PPL of $\sim80$, but this was achieved via dropout regularization \parencite{srivastava2014dropout}. There is a strong relationship between using dropout and training an ensemble of models. Thus, one can argue that a single model trained with dropout actually is not a single model, but an implicit ensemble (see also \cite{srivastava2014dropout}).  An ensemble of twenty simple RNNs and cache models did previously reach PPL as low as 72, while a single RNN model gives only 124 \parencite{mikolov_thesis}. \textcite{zaremba2014recurrent} trained an ensemble of 38 LSTMs regularized with dropout, each with ~100x times more parameters than the RNNs used by \cite{mikolov_thesis}, achieving PPL 68. This is arguably a small improvement over 72, and seems to strengthen our claim that dropout is an implicit model ensemble and thus should not be used when one wants to report the performance of a single model. However, the Delta-RNN is amenable to regularization, including drop-out. As our results show, when simple drop-out is applied, the Delta-RNN can reach much lower perplexities, even similar to the state-of-the-art with much larger models, especially when dynamic evaluation is permitted. This even extends to very complex architectures, such as the recently proposed TARDIS, which is a memory-augmented network (and when dynamic evaluation is used, the simple Delta-RNN can outperform this complex model). Though we investigate the utility of simple drop-out in this paper, our comparative results suggest that more sophisticated variants, such as variational drop-out \citep{gal2016theoretically}, could yield yet further improvement in performance.

\begin{figure*}
\begin{subfigure}{.95\textwidth}
  \centering
  \includegraphics[width=\linewidth]{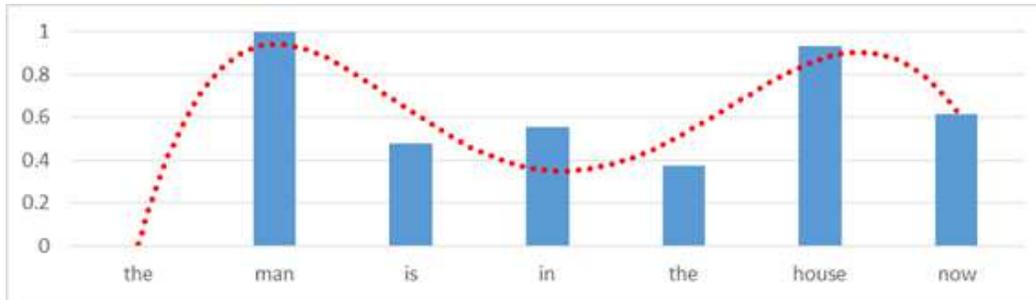}
  \label{sent1}
\end{subfigure}%
\vspace{0.005\textwidth}
\begin{subfigure}{.95\textwidth}
  \centering
  \includegraphics[width=\linewidth]{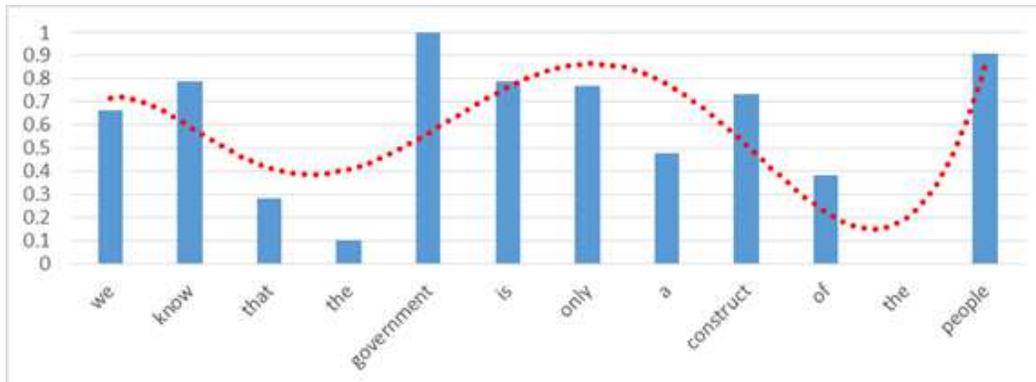}
  \label{sent2}
\end{subfigure}
\vspace{0.005\textwidth}
\begin{subfigure}{.95\textwidth}
  \centering
  \includegraphics[width=\linewidth]{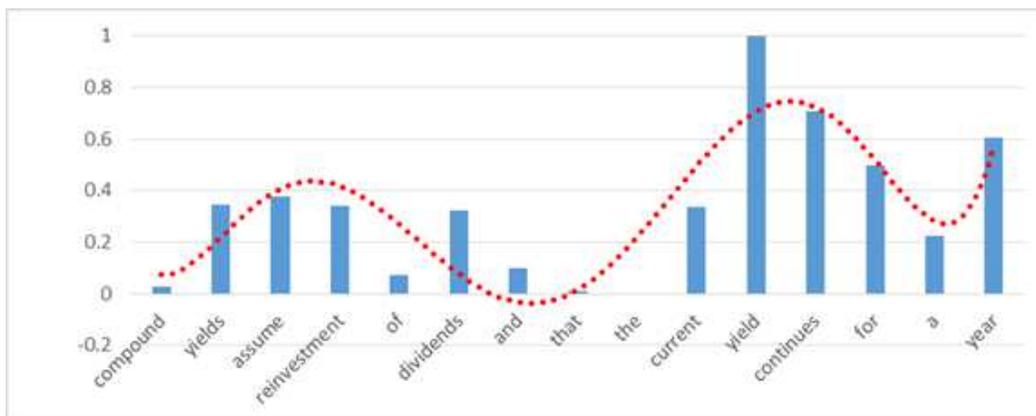}
  \label{sent3}
\end{subfigure}
\caption{L1 norm of deltas between consecutive states of model trained on Penn Treebank plotted over words of example sentences. A simple polynomial trend-line (dashed red) was fit to the bar heights in order to illustrate the informative `` bumps'' of each sample sentence. The main observation is that the norm is, in general, lower for low-information content words, such as the article ``the'', and higher for informative words, such as ``government''.}
\label{sent_samps}
\end{figure*}

What is the lesson to be learned from the Differential State Framework? First, and foremost, we can obtain strong performance in language modeling with a simpler, more efficient (in terms of number of parameters), and thus faster, architecture. Second, the Delta-RNN is designed from the interpretation that the computation of the next hidden state is the result of a composition of two functions. \textbf{One inner function decides how to ``propose'' a new hidden state while the outer function decides how to use this new proposal in updating the previously calculated state.} The data-driven interpolation mechanism is used by the model to decide how much impact the newly proposed state has in updating what is likely to be a slowly changing representation. The SCRN, which could be viewed as the predecessor to the Delta-RNN framework, was designed with the idea that some constrained units could serve as a sort of cache meant to capture longer-term dependencies. Like the SCRN, the Delta-RNN is designed to help mitigate the problem of vanishing gradients, and through the interpolation mechanism, has multiple pathways through which the gradient might be carried, boosting the error signal's longevity down the propagation path through time. However, the SCRN combines the slow-moving and fast-changing hidden states through a simple summation and thus cannot model non-linear interactions between its shorter and longer term memories, furthermore requiring tuning of the sizes of these separated layers. On the other hand, the Delta-RNN, which does not require special tuning of an additional hidden layer, can non-linearly combine the two types of states in a data-dependent fashion, possibly allowing the model to exploit boundary information from text, which is quite powerful in the case of documents. The key intuition is that the gating mechanism only allows the state proposal to affect the maintained memory state only if the currently observed data-point carries any useful information. This warrants a comparison, albeit indirect, to Surprisal Theory. This ``surprisal'' proves useful in iteratively forming a sentence impression that will help to better predict the words that come later.

With respect to the last point made, we briefly examine the evolution of a trained Delta-RNN's hidden state across several sample sentences.  The first two sentences are hand-created (constrained to use only the vocabulary of Penn Treebank) while the last one is sampled from the Penn Treebank training split. Since the Delta-RNN iteratively processes symbols of an ordered sequence, we measure the L1 norm across consecutive pairs of hidden states. We report the (min-max) normalized L1 scores\footnote{If we calculate the L1 norm, or Manhattan distance, for every contiguous pair of state vectors across a sequence of length $T$ and $\mathbf{h}_0$ is the state calculated for the start/null token, we obtain the sequence of L1 measures $L1_{seq} = \{L1_0(\mathbf{h}_0,\mathbf{h}_1),...,L1_{T}(\mathbf{h}_{T-1},\mathbf{h}_T)\}$ (the L1 for the start token is simply excluded). Calculating the score for any $\mathbf{h}_t$ ($t \in T$) is then as simple performing min-max normalization, or $L1_{score} = (L1(\mathbf{h}_{t-1},\mathbf{h}_t) - min(L1_{seq}))/(max(L1_{seq}) - min(L1_{seq}))$.} in Figure \ref{sent_samps} and observe that, in accordance with our intuition, we can see that the L1 norm is lower for high-frequency words (indicating a smaller delta) such as ``the'' or ``of'' or ``is'', which are words generally less informative about the general subject of a sentence/document. As this qualitative demonstration illustrates, the Delta-RNN appears to learn what to do with its internal state in the presence of symbols of variable information content.

\section{Conclusions}
We present the Differential State Framework, which affords us a useful perspective on viewing computation in recurrent neural networks. Instead of recomputing the whole state from scratch at every time step, the Delta-RNN only learns how to update the current state. This seems to be better suited for many types of problems, especially those that involve longer term patterns where part of the recurrent network's state should be constant most of the time. Comparison to the currently widely popular LSTM and GRU architectures shows that the Delta-RNN can achieve similar or better performance on language modeling tasks, while being conceptually much simpler and with far less parameters. Comparison to the Structurally Constrained Recurrent Network (SCRN), which shares many of the main ideas and motivation, shows better accuracy and a simpler model architecture (since, in the SCRN, tuning the sizes of two separate hidden layers is required, and this model cannot learn non-linear interactions within its longer memory).

Future work includes larger-scale language modeling experiments to test the efficacy of the Delta-RNN framework as well as architectural variants that employ decoupled memory. Since the Delta-RNN can also be stacked just as any other neural architecture, we intend to investigate if depth (in terms of hidden layers) might prove useful on larger-scale data-sets. In addition, we intend to explore how useful the Delta-RNN might be in other tasks that the architectures such as the LSTM currently hold state-of-the-art performance in. Finally, it would be useful to explore if Delta-RNN's simpler, faster design can speed up the performance of grander architectures, such as the Differentiable Neural Computer \parencite{graves2016hybrid} (composed of multiple LSTM modules).

\section*{Acknowledgments}
We thank C. Lee Giles and Prasenjit Mitra for their advice.  We thank NVIDIA for providing GPU hardware that supported this paper. A.O. was funded by a NACME-Sloan scholarship; D.R. acknowledges funding from NSF IIS-1459300.

\bibliographystyle{apa}  

\newpage
\input{NECO-03-17-2829R1-Supplementary_File}

\end{document}

%% file: NECO-03-17-2829R1-Supplementary_File.tex
\section*{Appendix A: Layer Normalized Delta-RNNs}
\label{model_details}
In this appendix, we describe how layer normalization would be applied to a Delta-RNN.  Though our preliminary experiments did not uncover that layer normalization gave much improvement over drop-out, this was only observed on the Penn Treebank benchmark.  Future work will investigate the benefits of layer normalization over drop-out (as well as model-ensembling ) on larger-scale benchmarks.

A simple RNN requires the layer normalization to be applied after calculating the full linear pre-activation (a sum of the filtration and the projected data point). On the other hand, a Delta-RNN requires further care (like the GRU) to ensure the correct components are normalized without damaging the favorable properties inherent to the model's multiplicative gating. If layer normalization is applied to the pre-activations of the late-integration Delta-RNN proposed in this paper, the update equations become:
\begin{align}
\mathbf{d}^{rec}_t &= LN(V_d \mathbf{h}_{t-1}),\ \mathbf{d}^{dat}_t = LN(W \mathbf{x}_t)\\
\mathbf{z}_{t} &= \phi_{hid}( \mathbf{d}^{rec}_t \otimes \mathbf{d}^{dat}_t +  \mathbf{d}^{rec}_t + \mathbf{d}^{dat}_t)\mbox{,  and,} \\
\mathbf{h}_{t} &=  \Phi( (1 - \mathbf{r}) \otimes \mathbf{z}_{t} + \mathbf{r} \otimes \mathbf{h}_{t-1} ) \mbox{,  and,} \\ \mathbf{r} &=  1 / (1 + exp(-[\mathbf{d}^{dat}_t + \mathbf{b}_r])) \mbox{.}
\end{align}
Note that the additional bias parameters introduced in the original update equations are now omitted. This can be done since the layer normalization operation applied will now perform the work of shifting and scaling. Since the Delta-RNN takes advantage of parameter-sharing, it notably requires substantially fewer layer normalizations than a more complex model (such as the GRU) would. A standard GRU would require nine layer normalizations while the Delta-RNN simply requires two.

%% file: NECO-03-17-2829R1-Source.bbl
\begin{thebibliography}{}

\bibitem[\protect\astroncite{Aslin et~al.}{1998}]{aslin1998infants}
Aslin, R.~N., Saffran, J.~R., and Newport, E.~L. (1998).
\newblock Computation of conditional probability statistics by 8-month-old
  infants.
\newblock {\em Psychological Science}, 9(4):321--324.

\bibitem[\protect\astroncite{Ba et~al.}{2016}]{ba2016layer}
Ba, J.~L., Kiros, J.~R., and Hinton, G.~E. (2016).
\newblock Layer normalization.
\newblock {\em arXiv preprint arXiv:1607.06450}.

\bibitem[\protect\astroncite{Baayen and Schreuder}{2006}]{baayen2006morph}
Baayen, R.~H. and Schreuder, R. (2006).
\newblock {\em Morphological Processing}.
\newblock Wiley.

\bibitem[\protect\astroncite{Bahdanau et~al.}{2014}]{bahdanau2014neural}
Bahdanau, D., Cho, K., and Bengio, Y. (2014).
\newblock Neural machine translation by jointly learning to align and
  translate.
\newblock {\em arXiv preprint arXiv:1409.0473}.

\bibitem[\protect\astroncite{Boston et~al.}{2008}]{boston2008parsingcosts}
Boston, M.~F., Hale, J., Kliegl, R., Patil, U., and Vasishth, S. (2008).
\newblock Parsing costs as predictors of reading difficulty: An evaluation
  using the {Potsdam Sentence Corpus}.
\newblock {\em Journal of Eye Movement Research}, 2(1).

\bibitem[\protect\astroncite{Choudhury}{2015}]{choudhury_thought_2015}
Choudhury, V. (2015).
\newblock Thought vectors: Bringing common sense to artificial intelligence.
\newblock www.iamwire.com.

\bibitem[\protect\astroncite{Chung et~al.}{2016}]{chung2016hierarchical}
Chung, J., Ahn, S., and Bengio, Y. (2016).
\newblock Hierarchical multiscale recurrent neural networks.
\newblock {\em arXiv preprint arXiv:1609.01704}.

\bibitem[\protect\astroncite{Chung et~al.}{2014}]{chung2014empirical}
Chung, J., Gulcehre, C., Cho, K., and Bengio, Y. (2014).
\newblock Empirical evaluation of gated recurrent neural networks on sequence
  modeling.
\newblock {\em arXiv preprint arXiv:1412.3555}.

\bibitem[\protect\astroncite{Chung et~al.}{2015}]{chung2015gated}
Chung, J., Gulcehre, C., Cho, K., and Bengio, Y. (2015).
\newblock Gated feedback recurrent neural networks.
\newblock In {\em International Conference on Machine Learning}, pages
  2067--2075.

\bibitem[\protect\astroncite{Cooijmans et~al.}{2016}]{cooijmans2016recurrent}
Cooijmans, T., Ballas, N., Laurent, C., G{\"u}l{\c{c}}ehre, {\c{C}}., and
  Courville, A. (2016).
\newblock Recurrent batch normalization.
\newblock {\em arXiv preprint arXiv:1603.09025}.

\bibitem[\protect\astroncite{Das et~al.}{1992}]{das1992learning}
Das, S., Giles, C.~L., and Sun, G.-Z. (1992).
\newblock Learning context-free grammars: Capabilities and limitations of a
  recurrent neural network with an external stack memory.
\newblock In {\em Proceedings of the 14th Annual Conference of the {Cognitive
  Science Society}}, page~14, Bloomington, IN.

\bibitem[\protect\astroncite{Elman}{1990}]{elman1990finding}
Elman, J.~L. (1990).
\newblock Finding structure in time.
\newblock {\em Cognitive Science}, 14(2):179--211.

\bibitem[\protect\astroncite{Gal and Ghahramani}{2016}]{gal2016theoretically}
Gal, Y. and Ghahramani, Z. (2016).
\newblock A theoretically grounded application of dropout in recurrent neural
  networks.
\newblock In {\em Advances in Neural Information Processing Systems}, pages
  1019--1027.

\bibitem[\protect\astroncite{Gers and Schmidhuber}{2000}]{gers2000recurrent}
Gers, F.~A. and Schmidhuber, J. (2000).
\newblock Recurrent nets that time and count.
\newblock In {\em Proceedings of the {IEEE-INNS-ENNS} International Joint
  Conference on Neural Networks}, volume~3, pages 189--194. IEEE.

\bibitem[\protect\astroncite{Giles et~al.}{1991}]{giles1991second}
Giles, C.~L., Chen, D., Miller, C., Chen, H., Sun, G., and Lee, Y. (1991).
\newblock Second-order recurrent neural networks for grammatical inference.
\newblock In {\em International Joint Conference on Neural Networks}, volume~2,
  pages 273--281.

\bibitem[\protect\astroncite{Giles et~al.}{2001}]{giles2001noisy}
Giles, C.~L., Lawrence, S., and Tsoi, A.~C. (2001).
\newblock Noisy time series prediction using recurrent neural networks and
  grammatical inference.
\newblock {\em Machine Learning}, 44(1-2):161--183.

\bibitem[\protect\astroncite{Giles et~al.}{1992}]{giles1992learning}
Giles, C.~L., Miller, C.~B., Chen, D., Chen, H.-H., Sun, G.-Z., and Lee, Y.-C.
  (1992).
\newblock Learning and extracting finite state automata with second-order
  recurrent neural networks.
\newblock {\em Neural Computation}, 4(3):393--405.

\bibitem[\protect\astroncite{Goudreau et~al.}{1994}]{goudreau1994first}
Goudreau, M.~W., Giles, C.~L., Chakradhar, S.~T., and Chen, D. (1994).
\newblock First-order versus second-order single-layer recurrent neural
  networks.
\newblock {\em IEEE Transactions on Neural Networks}, 5(3):511--513.

\bibitem[\protect\astroncite{Graves}{2013}]{graves2013generating}
Graves, A. (2013).
\newblock Generating sequences with recurrent neural networks.
\newblock {\em arXiv preprint arXiv:1308.0850}.

\bibitem[\protect\astroncite{Graves et~al.}{2016}]{graves2016hybrid}
Graves, A., Wayne, G., Reynolds, M., Harley, T., Danihelka, I.,
  Grabska-Barwi{\'n}ska, A., Colmenarejo, S.~G., Grefenstette, E., Ramalho, T.,
  Agapiou, J., et~al. (2016).
\newblock Hybrid computing using a neural network with dynamic external memory.
\newblock {\em Nature}, 538(7626):471--476.

\bibitem[\protect\astroncite{Gulcehre et~al.}{2016}]{gulcehre2016noisy}
Gulcehre, C., Moczulski, M., Denil, M., and Bengio, Y. (2016).
\newblock Noisy activation functions.
\newblock {\em arXiv preprint arXiv:1603.00391}.

\bibitem[\protect\astroncite{Ha et~al.}{2016}]{ha2016hypernetworks}
Ha, D., Dai, A., and Le, Q.~V. (2016).
\newblock Hypernetworks.
\newblock {\em arXiv preprint arXiv:1609.09106}.

\bibitem[\protect\astroncite{Hale}{2001}]{hale2001surprisal}
Hale, J. (2001).
\newblock A probabilistic {Earley} parser as a psycholinguistic model.
\newblock In {\em Proceedings of the Second Meeting of the North American
  Chapter of the {Association for Computational Linguistics}}, NAACL '01, pages
  1--8, Stroudsburg, PA, USA.

\bibitem[\protect\astroncite{He et~al.}{2016}]{he2016identity}
He, K., Zhang, X., Ren, S., and Sun, J. (2016).
\newblock Identity mappings in deep residual networks.
\newblock In {\em European Conference on Computer Vision}, pages 630--645.
  Springer.

\bibitem[\protect\astroncite{Hochreiter and
  Schmidhuber}{1997a}]{hochreiter1997long}
Hochreiter, S. and Schmidhuber, J. (1997a).
\newblock Long short-term memory.
\newblock {\em Neural Computation}, 9(8):1735--1780.

\bibitem[\protect\astroncite{Hochreiter and
  Schmidhuber}{1997b}]{hochreiter1997lstm}
Hochreiter, S. and Schmidhuber, J. (1997b).
\newblock {LTSM} can solve hard time lag problems.
\newblock In {\em Advances in {Neural Information Processing Systems}:
  {P}roceedings of the 1996 Conference}, pages 473--479.

\bibitem[\protect\astroncite{Ioffe and Szegedy}{2015}]{ioffe2015batch}
Ioffe, S. and Szegedy, C. (2015).
\newblock Batch normalization: Accelerating deep network training by reducing
  internal covariate shift.
\newblock {\em arXiv preprint arXiv:1502.03167}.

\bibitem[\protect\astroncite{Jernite et~al.}{2016}]{jernite2016variable}
Jernite, Y., Grave, E., Joulin, A., and Mikolov, T. (2016).
\newblock Variable computation in recurrent neural networks.
\newblock {\em arXiv preprint arXiv:1611.06188}.

\bibitem[\protect\astroncite{Jordan}{1990}]{jordan1986attractor}
Jordan, M.~I. (1990).
\newblock Artificial neural networks.
\newblock chapter Attractor Dynamics and Parallelism in a Connectionist
  Sequential Machine, pages 112--127. IEEE Press, Piscataway, NJ, USA.

\bibitem[\protect\astroncite{Joulin and Mikolov}{2015}]{joulin2015inferring}
Joulin, A. and Mikolov, T. (2015).
\newblock Inferring algorithmic patterns with stack-augmented recurrent nets.
\newblock In {\em Advances in {Neural Information Processing Systems}}, pages
  190--198.

\bibitem[\protect\astroncite{Jozefowicz et~al.}{2015}]{zaremba2015empirical}
Jozefowicz, R., Zaremba, W., and Sutskever, I. (2015).
\newblock An empirical exploration of recurrent network architectures.
\newblock In {\em Proceedings of the 32nd International Conference on Machine
  Learning (ICML-15)}, pages 2342--2350.

\bibitem[\protect\astroncite{Kingma and Ba}{2014}]{kingma2014adam}
Kingma, D. and Ba, J. (2014).
\newblock Adam: A method for stochastic optimization.
\newblock {\em arXiv preprint arXiv:1412.6980}.

\bibitem[\protect\astroncite{Koutnik et~al.}{2014}]{koutnik2014clockwork}
Koutnik, J., Greff, K., Gomez, F., and Schmidhuber, J. (2014).
\newblock A clockwork rnn.
\newblock {\em arXiv preprint arXiv:1402.3511}.

\bibitem[\protect\astroncite{Krueger et~al.}{2016}]{krueger2016zoneout}
Krueger, D., Maharaj, T., Kram{\'a}r, J., Pezeshki, M., Ballas, N., Ke, N.~R.,
  Goyal, A., Bengio, Y., Larochelle, H., Courville, A., et~al. (2016).
\newblock Zoneout: Regularizing rnns by randomly preserving hidden activations.
\newblock {\em arXiv preprint arXiv:1606.01305}.

\bibitem[\protect\astroncite{Le et~al.}{2015}]{le2015simple}
Le, Q.~V., Jaitly, N., and Hinton, G.~E. (2015).
\newblock A simple way to initialize recurrent networks of rectified linear
  units.
\newblock {\em arXiv preprint arXiv:1504.00941}.

\bibitem[\protect\astroncite{Levy}{2008}]{levy2008expectation}
Levy, R. (2008).
\newblock Expectation-based syntactic comprehension.
\newblock {\em Cognition}, 106(3):1126 -- 1177.

\bibitem[\protect\astroncite{Maas et~al.}{2011}]{maas_2011_ACL_HLT2011}
Maas, A.~L., Daly, R.~E., Pham, P.~T., Huang, D., Ng, A.~Y., and Potts, C.
  (2011).
\newblock Learning word vectors for sentiment analysis.
\newblock In {\em Proceedings of the 49th Annual Meeting of the {Association
  for Computational Linguistics}: {Human Language Technologies}}, ACL-HLT2011,
  pages 142--150, Portland, Oregon, USA. Association for Computational
  Linguistics.

\bibitem[\protect\astroncite{Marcus et~al.}{1993}]{marcus1993building}
Marcus, M.~P., Marcinkiewicz, M.~A., and Santorini, B. (1993).
\newblock Building a large annotated corpus of {English}: The {Penn Treebank}.
\newblock {\em Computational Linguistics}, 19(2):313--330.

\bibitem[\protect\astroncite{Mikolov}{2012}]{mikolov_thesis}
Mikolov, T. (2012).
\newblock {\em Statistical Language Models Based on Neural Networks}.
\newblock PhD thesis, University of Brno, Brno, CZ.

\bibitem[\protect\astroncite{Mikolov et~al.}{2014}]{mikolov2014learning}
Mikolov, T., Joulin, A., Chopra, S., Mathieu, M., and Ranzato, M. (2014).
\newblock Learning longer memory in recurrent neural networks.
\newblock {\em arXiv preprint arXiv:1412.7753}.

\bibitem[\protect\astroncite{Mikolov et~al.}{2010}]{mikolov2010recurrent}
Mikolov, T., Karafi{\'a}t, M., Burget, L., \v{C}ernock{\'y}, J., and Khudanpur,
  S. (2010).
\newblock Recurrent neural network based language model.
\newblock In {\em Proceedings of the 11th Annual Conference of the
  International Speech Communication Association (INTERSPEECH 2010)}, volume~2,
  pages 1045--1048, Makuhari, Chiba, JP.

\bibitem[\protect\astroncite{Mikolov et~al.}{2011}]{mikolov2011extensions}
Mikolov, T., Kombrink, S., Burget, L., {\v{C}}ernock{\'y}, J., and Khudanpur,
  S. (2011).
\newblock Extensions of recurrent neural network language model.
\newblock In {\em 2011 {IEEE International Conference on Acoustics, Speech and
  Signal Processing (ICASSP)}}, pages 5528--5531, Prague, Czech Republic.

\bibitem[\protect\astroncite{Mikolov et~al.}{2012}]{mikolov2012subword}
Mikolov, T., Sutskever, I., Deoras, A., Le, H.-S., Kombrink, S., and
  {\v{C}}ernock{\'y}, J. (2012).
\newblock Subword language modeling with neural networks.
\newblock \url{http://www.fit.vutbr.cz/\~imikolov/rnnlm/char.pdf}.
\newblock Accessed: 2017-06-01.

\bibitem[\protect\astroncite{Mozer}{1993}]{mozer1993neural}
Mozer, M.~C. (1993).
\newblock Neural net architectures for temporal sequence processing.
\newblock In {\em Santa Fe Institute Studies in the Sciences of Complexity},
  volume~15, pages 243--243. Addison-Wesley Publishing Co.

\bibitem[\protect\astroncite{Neal}{2012}]{neal2012bayesian}
Neal, R.~M. (2012).
\newblock {\em Bayesian learning for neural networks}, volume 118.
\newblock Springer Science \& Business Media.

\bibitem[\protect\astroncite{Pascanu et~al.}{2013}]{pascanu2013difficulty}
Pascanu, R., Mikolov, T., and Bengio, Y. (2013).
\newblock On the difficulty of training recurrent neural networks.
\newblock {\em International Conference of Machine Learning (3)},
  28:1310--1318.

\bibitem[\protect\astroncite{Polyak and
  Juditsky}{1992}]{polyak1992acceleration}
Polyak, B.~T. and Juditsky, A.~B. (1992).
\newblock Acceleration of stochastic approximation by averaging.
\newblock {\em SIAM Journal on Control and Optimization}, 30(4):838--855.

\bibitem[\protect\astroncite{Serban et~al.}{2016}]{serban2016multi}
Serban, I.~V., Ororbia, I., Alexander, G., Pineau, J., and Courville, A.
  (2016).
\newblock Piecewise Latent Variables for Neural Variational Text Processing.
\newblock {\em arXiv preprint arXiv:1612.00377}.

\bibitem[\protect\astroncite{Srivastava et~al.}{2014}]{srivastava2014dropout}
Srivastava, N., Hinton, G.~E., Krizhevsky, A., Sutskever, I., and
  Salakhutdinov, R. (2014).
\newblock Dropout: a simple way to prevent neural networks from overfitting.
\newblock {\em Journal of Machine Learning Research}, 15(1):1929--1958.

\bibitem[\protect\astroncite{Sukhbaatar
  et~al.}{2015}]{sukhbaatar_end_to_end_2015}
Sukhbaatar, S., Szlam, A., Weston, J., and Fergus, R. (2015).
\newblock End-to-end memory networks.
\newblock {\em {arXiv}:1503.08895 [cs]}.

\bibitem[\protect\astroncite{Sun et~al.}{1998}]{sun1998neural}
Sun, G.-Z., Giles, C.~L., and Chen, H.-H. (1998).
\newblock The neural network pushdown automaton: Architecture, dynamics and
  training.
\newblock In {\em Adaptive processing of sequences and data structures}, pages
  296--345. Springer.

\bibitem[\protect\astroncite{Sundermeyer}{2016}]{sundermeyer2016}
Sundermeyer, M. (2016).
\newblock {\em Improvements in Language and Translation Modeling}.
\newblock PhD thesis, RWTH Aachen University.

\bibitem[\protect\astroncite{Turian et~al.}{2009}]{turian2009quadratic}
Turian, J., Bergstra, J., and Bengio, Y. (2009).
\newblock Quadratic features and deep architectures for chunking.
\newblock In {\em Proceedings of {Human Language Technologies}: The 2009 Annual
  Conference of the North American Chapter of the {Association for
  Computational Linguistics}, Companion Volume: Short Papers}, pages 245--248.
  Association for Computational Linguistics.

\bibitem[\protect\astroncite{Wang and Cho}{2015}]{wang2015larger}
Wang, T. and Cho, K. (2015).
\newblock Larger-context language modelling.
\newblock {\em arXiv preprint arXiv:1511.03729}.

\bibitem[\protect\astroncite{Weston et~al.}{2014}]{weston_memory_2014}
Weston, J., Chopra, S., and Bordes, A. (2014).
\newblock Memory networks.
\newblock {\em {arXiv}:1410.3916 [cs, stat]}.

\bibitem[\protect\astroncite{Wu et~al.}{2016}]{wu_multiplicative_2016}
Wu, Y., Zhang, S., Zhang, Y., Bengio, Y., and Salakhutdinov, R.~R. (2016).
\newblock On multiplicative integration with recurrent neural networks.
\newblock In {\em Advances in Neural Information Processing Systems}, pages
  2856--2864.
  
\bibitem[\protect\astroncite{Gulcehre et~al.}{2017}]{gulcehre2017memory}
Gulcehre, Caglar and Chandar, Sarath and Bengio, Yoshua (2017).
\newblock Memory Augmented Neural Networks with Wormhole Connections.
\newblock {\em {arXiv}:1701.08718 [cs, stat]}.
\bibitem[\protect\astroncite{Zaremba et~al.}{2014}]{zaremba2014recurrent}
Zaremba, W., Sutskever, I., and Vinyals, O. (2014).
\newblock Recurrent neural network regularization.
\newblock {\em arXiv preprint arXiv:1409.2329}.

\bibitem[\protect\astroncite{Zhou et~al.}{2016}]{zhou2016minimal}
Zhou, G.-B., Wu, J., Zhang, C.-L., and Zhou, Z.-H. (2016).
\newblock Minimal gated unit for recurrent neural networks.
\newblock {\em International Journal of Automation and Computing},
  13(3):226--234.

\end{thebibliography}
